\documentclass[letterpaper, 10 pt, conference]{ieeeconf}

\IEEEoverridecommandlockouts
\usepackage{cite}
\usepackage{amsmath,amssymb,amsfonts}
\usepackage{algorithmic}
\usepackage{graphicx}
\usepackage{textcomp}
\usepackage{xcolor}
\def\BibTeX{{\rm B\kern-.05em{\sc i\kern-.025em b}\kern-.08em
    T\kern-.1667em\lower.7ex\hbox{E}\kern-.125emX}}

\usepackage{soul, color}
\usepackage{colortbl}
\definecolor{Gray}{gray}{0.9}
    
\begin{document}

\title{Recognition of Dynamic Hand Gestures in Long Distance using a Web-Camera for Robot Guidance}


\author{Eran Bamani Beeri, Eden Nissinman and Avishai Sintov
\thanks{E. Bamani Beeri, E. Nissinman and A. Sintov are with the School of Mechanical Engineering, Tel-Aviv University, Israel. E-mail: eranbamani@mail.tau.ac.il,  edennissinman@gmail.com, sintov1@tauex.tau.ac.il}
\thanks{This work was supported by the Israel Innovation Authority (grant No. 77857).} 
}

\maketitle
\thispagestyle{empty}
\pagestyle{empty}

\begin{abstract}
Dynamic gestures enable the transfer of directive information to a robot. Moreover, the ability of a robot to recognize them from a long distance makes communication more effective and practical. However, current state-of-the-art models for dynamic gestures exhibit limitations in recognition distance, typically achieving effective performance only within a few meters. In this work, we propose a model for recognizing dynamic gestures from a long distance of up to 20 meters. The model integrates the SlowFast and Transformer architectures (SFT) to effectively process and classify complex gesture sequences captured in video frames. SFT demonstrates superior performance over existing models.
\end{abstract}



\section{Introduction}

In Human-Robot Interaction (HRI), gesture recognition is imperative to deliver information, eliminate complex interfaces, and improve user experience \cite{Shanthakumar, Zheng2023}. With gestures, a user can convey nonverbal and simple commands even from a long distance without the need to shout. For instance, a user may direct robot movements with simple pointing gestures \cite{bamani2023recognition}. However, most gesture recognition approaches enable short-range interactions for up to a few meters \cite{nickel2007visual, wachs2011vision}. Nevertheless, the authors have recently proposed a static gesture recognition model using a web camera with an effective distance of up to at least 25 meters while also facing challenges of low-resolution and occlusions \cite{bamani2024ultra}.

The majority of gesture recognition approaches consider only static gestures. However, some gestures are inherently dynamic such as waving and beckoning. Additionally, some dynamic gestures may appear as different gestures if only a single frame is observed. For example, a snapshot of a waving gesture may look like a stop gesture. State-of-the-art dynamic gesture recognition considers only close-range gesture recognition \cite{Yu2022}.

\begin{figure}[htbp]
\centering
\includegraphics[width=\linewidth]{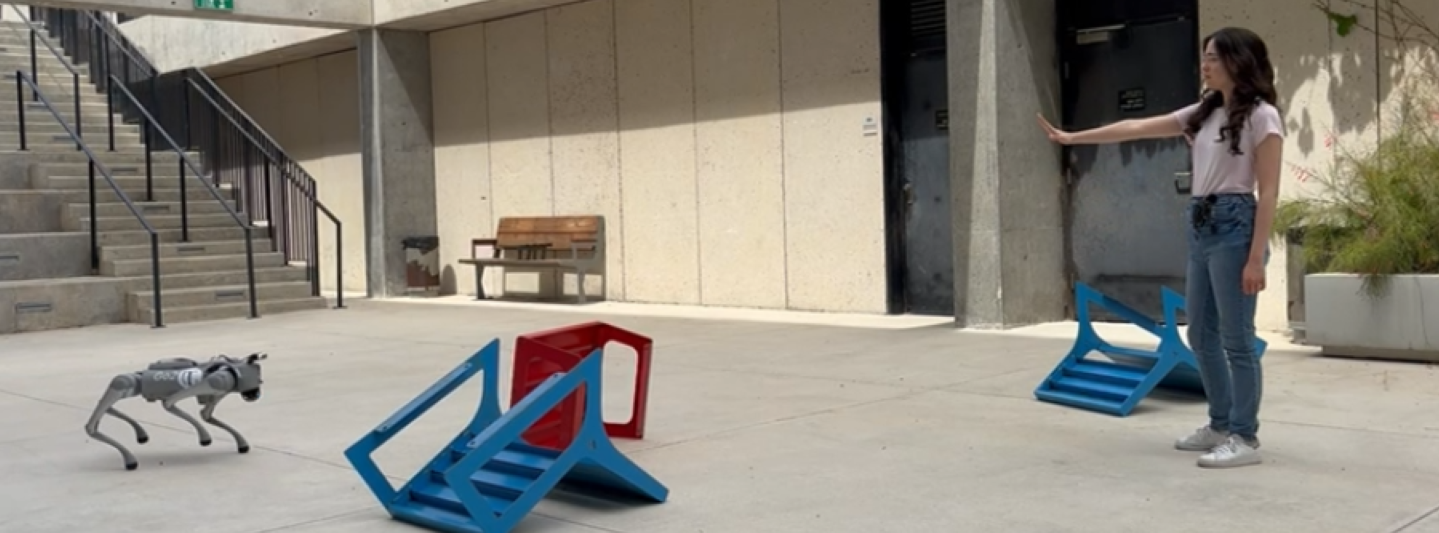}
\vspace{-0.5cm}
\caption{Directing a robot through obstacles using gestures.}
\label{fig:objavoid}
\vspace{-0.5cm}
\end{figure}

In this work, we address the problem of dynamic gesture recognition in the long range. Specifically, we propose the Slow-Fast-Transformer (SFT) model based on the SlowFast \cite{Feichtenhofer} and Transformer architectures \cite{Vaswani}. Also, a specialized loss function for long-range recognition is proposed. The approach can then be used to direct a robot to perform desired tasks or bypass obstacles as seen in Fig. \ref{fig:objavoid}.




\section{Methods}
\subsection{Problem Formulation}

We consider a set of $m$ gestures $\{\mathcal{O}_1, \ldots, \mathcal{O}_m\}$ where each can be either static or dynamic. A gesture can be exhibited within a distance $d\leq20~m$ from the camera. The gesture recognition model will solve the following problem:
\begin{equation}
    j^* = \arg\max_j P(\mathcal{O}_j \mid \mathbf{I}_{i-K},\ldots,\mathbf{I}_i), \quad \forall j = 1, \ldots, m
\end{equation}
where $P(\mathcal{O}_j \mid \mathbf{I}_{i-K},\ldots,\mathbf{I}_i)$ is the probability that the exhibited gesture is $\mathcal{O}_j$ given a set of $K+1$ past images $\mathbf{I}_{i-K},\ldots,\mathbf{I}_i$.



\subsection{Data Collection and Pre-processing}

We collected a dataset $\mathcal{D}$ of continuous videos capturing human gestures. Let $V_i=\{\mathbf{I}_1,\ldots,\mathbf{I}_n\}$ represent a video clip of a dynamic gesture $o_i$ taken with distance $d_i$ from the camera. The video is constructed of $n$ frames where each frame $\mathbf{I}_j$ contains instantaneous visual information from the camera. Consequently, a collection of $N$ videos yields labeled dataset $\mathcal{D} = \{(V_i, d_i, o_i)\}_{i=1}^{N}$.

In the next step, the videos in the dataset are pre-processed to reduce video frames and crop out objects to streamline the search area and minimize computational effort. Given a video $V_i$ with $n$ frames, we reduce the number of frames to $k < n$ using K-Means clustering. Each frame $\mathbf{I}_j\in V_i$ is processed through a pre-trained ResNet \cite{He} to extract a low-dimensional feature vector $\mathbf{v}_j \in \mathbb{R}^d$. All feature vectors associated with $V_i$ are clustered into $k$ clusters. The frame corresponding to the feature vector closest to its cluster's centroid is selected as the representative frame. This leads to $k$ representative frames $\{\mathbf{I}_{rep,1},\ldots,\mathbf{I}_{rep,k}\}$. YOLOv3 \cite{Redmon} is then used to detect the user in the frame and crop out the background. This step enhances focus particularly when the user is far from the camera. Finally, the cropped image is resized to $224 \times 224$ to ensure uniformity across the dataset, maintaining a consistent size and ratio for the model.






\subsection{Model}

We propose the integration of the SlowFast network with a Transformer for recognizing dynamic gestures in the long range. The SlowFast Network architecture processes video frames at multiple temporal resolutions to capture slow and fast motion dynamics. An inputted video clip 
goes through two pathways. The slow and fast pathways of SlowFast handle lower and higher temporal resolutions, respectively. Their outputs are concatenated into a combined feature representation. The Transformer model processes SlowFast features with convolutional layers, an embedding layer, and four transformer encoders. The output is temporally pooled and fed into a fully connected layer for class scores. 

We propose a distance-weighted cross-entropy loss function termed \textit{LongLoss} for training the SFT model while accounting for user distance. Given predicted output $\tilde{\mathbf{o}}_i$, target label $\mathbf{o}_i$ and distance $\mathbf{d}_i$, the loss is given by
\vspace{-0.2cm}
\begin{equation}
\mathcal{L} = \frac{1}{B} \sum_{i=1}^{B} \left[ \text{CE}(\tilde{\mathbf{o}}_i, \mathbf{o}_i) \times \left(1 + \frac{\alpha (d_i - b_0)}{b_1 - b_0} \right) \right]
\vspace{-0.2cm}
\end{equation}
where $B$ is the batch size, $\alpha$ is a distance weighting factor, $b_0$ and $b_1$ are predefined thresholds, and $\text{CE}(\cdot)$ is the cross-entropy loss function.







\section{Experimental Results}

\begin{figure}
    \centering
    \begin{tabular}{ccccc}
        \includegraphics[width=0.16\linewidth]{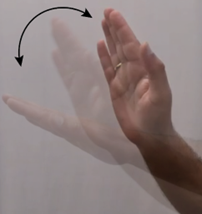} &
        \includegraphics[width=0.16\linewidth]{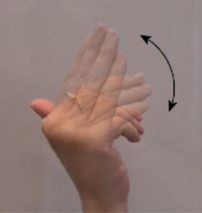} &
        \includegraphics[width=0.16\linewidth]{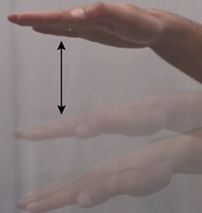} &
        \includegraphics[width=0.16\linewidth]{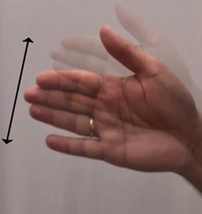} &
        \includegraphics[width=0.16\linewidth]{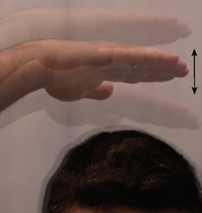} \vspace{-0.1cm}\\
        \small{(a)} & \small{(b)} & \small{(c)} & \small{(d)} & \small{(e)} \\
    \end{tabular}
    
    
    \begin{tabular}{cccc}
        \includegraphics[width=0.16\linewidth]{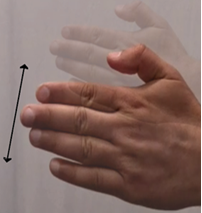} &
        \includegraphics[width=0.16\linewidth]{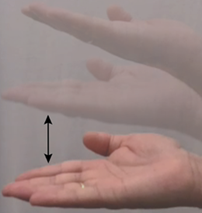} &
        \includegraphics[width=0.16\linewidth]{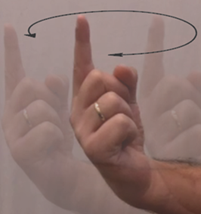} &
        \includegraphics[width=0.16\linewidth]{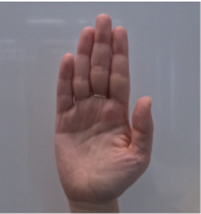} \\
        \small{(f)} & \small{(g)} & \small{(h)} & \small{(i)} \vspace{-0.1cm}\\
    \end{tabular}
    
    \caption{Nine hand gestures used in this study.}
    \label{fig:human_gestures}
    \vspace{-0.7cm}
\end{figure}

To assess the effectiveness of the proposed model, we conducted a series of experiments comparing its performance with that of state-of-the-art video recognition models. We focus on ten human gestures where nine are seen in Figure \ref{fig:human_gestures} and include: (a) go back; (b) beckoning; (c) lower; (d) move left; (e) follow me; (f) move right; (g) higher; (h) spin; and (i) stop. The last gesture is null. 

\begin{figure}[htbp]
\centering
\includegraphics[width=\linewidth]{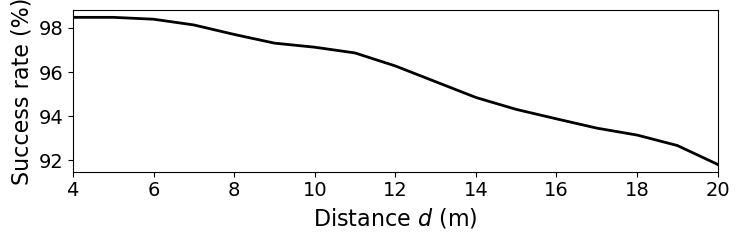}
\vspace{-0.8cm}
\caption{Gesture recognition success rate of the SFT model with regard to the distance $d$ of the user from the camera.}
\label{fig:performance_vs_distance}
\vspace{-0.5cm}
\end{figure}

The collected dataset $\mathcal{D}$ included $N=3,240$ video samples of hand gestures observed from 4 to 20 meters in indoor and outdoor environments. Within each interval of one meter, 20 video samples of 4 seconds were recorded. In addition, a set of 320 labeled videos was recorded for a test set in entirely different environments. Each video sample has $n \leq 84$ frames later reduced to $k = 8$. SFT was trained using $\mathcal{D}$. The dataset was also used to train other video recognition models including Swin \cite{Liu}, ViViT \cite{Arnab}, SlowFast, and X3D \cite{Feichtenhofer2} for comparison. SFT is also evaluated with standard CE loss. Recognition success rate with respect to the test set is reported in Table \ref{tab1} along with loss and Mean Average Precision (mAP). The results clearly show significant performance improvement using SFT over the other models. Figure \ref{fig:performance_vs_distance} shows the recognition success rate with respect to the distance from the camera.


\begin{table}[htbp]
\caption{Comparison of Video Recognition Models}
\vspace{-0.5cm}
\begin{center}
\begin{tabular}{lccc}
\hline
{Model} & {Accuracy (\%)} & {Loss} & {mAP} \\
\hline
 Swin \cite{Liu} & 80.5 & 0.41 & 0.76 \\
 ViViT \cite{Arnab} & 78.3 & 0.47 & 0.71 \\
 SlowFast  & 75.4 & 0.46 & 0.69 \\
X3D \cite{Feichtenhofer2} & 81.2 & 0.39 &  0.78 \\
SFT w/ CE loss & 89.1 & 0.25 & 0.86 \\
\rowcolor{Gray}
SFT & 95.7 & 0.13 & 0.93  \\
\hline
\end{tabular}
\label{tab1}
\end{center}
\vspace{-0.3cm}
\end{table}


\section{Conclusion}
This paper presents a model for recognizing dynamic gestures from a long distance of up to 20 meters in various environments. Such a distance is practical for the natural human guidance of a robot. Combining advanced video recognition techniques with distance-aware classification, our approach demonstrates superior performance. Future work could focus on expanding gesture vocabulary and real-time optimization.


\bibliographystyle{IEEEtran}
\bibliography{ref}

\end{document}